\documentclass[conference]{IEEEtran}

\usepackage[a4paper, top=1.9cm, bottom=2.54cm, left=1.9cm, right=1.9cm]{geometry}

\usepackage{amsmath}     

\usepackage{graphicx}    
\usepackage{booktabs}
\usepackage{url}
\usepackage[noadjust]{cite}

\usepackage[pdftex]{hyperref}
\hypersetup{
    colorlinks=true,
    linkcolor=blue,
    filecolor=magenta,
    urlcolor=blue,
    citecolor=blue,
}

\begin{document}

\title{oboro: Text-to-Image Synthesis on Limited Data using Flow-based Diffusion Transformer with MMH Attention}

\author{
  \IEEEauthorblockN{
   Ryusuke Mizutani\IEEEauthorrefmark{1},
   Kazuaki Matano\IEEEauthorrefmark{1}\IEEEauthorrefmark{2},
   Tsugumi Kadowaki\IEEEauthorrefmark{1},
   Haruki Tenya\IEEEauthorrefmark{1},
   Layris\IEEEauthorrefmark{1},
   nuigurumi\IEEEauthorrefmark{1},\\
   Koki Hashimoto\IEEEauthorrefmark{1},
   Yu Tanaka\IEEEauthorrefmark{1}
  }

  \IEEEauthorblockA{
    \IEEEauthorrefmark{1}\textit{AiHUB Inc., Tokyo 101-0041, Japan}
  }Email: oboro@aihub.tokyo
  \IEEEauthorblockA{
    \IEEEauthorrefmark{2}\textit{Department of Physics, Okayama University of Science, Okayama 700-0005, Japan}}
}

\maketitle

\begin{abstract}
This project was conducted as a 2nd-term adopted project of the ``Post-5G Information and Communication System Infrastructure Enhancement R\&D Project / Development of Competitive Generative AI Foundation Models (GENIAC)," a business of the Ministry of Economy, Trade and Industry (METI) and the New Energy and Industrial Technology Development Organization (NEDO). To address challenges such as labor shortages in Japan's anime production industry, this project aims to develop an image generation model from scratch. This report details the technical specifications of the developed image generation model, ``oboro:." We have developed ``oboro:," a new image generation model built from scratch, using only copyright-cleared images for training. A key characteristic is its architecture, designed to generate high-quality images even from limited datasets. The foundation model weights and inference code are publicly available alongside this report. This project marks the first release of an open-source, commercially-oriented image generation AI fully developed in Japan. AiHUB originated from the OSS community; by maintaining transparency in our development process, we aim to contribute to Japan's AI researcher and engineer community and promote the domestic AI development ecosystem.
\end{abstract}

\begin{IEEEkeywords}
Generative AI, Image Generation, Diffusion Models, Diffusion Transformer (DiT), GENIAC, oboro, Anime Production, Copyright
\end{IEEEkeywords}

\section{Development Background}
The release of Stable Diffusion (SD) \cite{rombach2022highres} in August 2022 was a turning point in the history of Text-to-Image (T2I) generation AI. High-performance T2I generation technology, previously limited to a small number of researchers and companies, was released as open source. This accelerated development within the open-source software (OSS) community and created ripple effects in business applications.

Although early models produced outputs with noticeable artifacts, the technology has evolved rapidly. While image generation models are already being used in various domains, their rapid maturation ahead of established legal frameworks and social consensus has created numerous challenges.

\subsection{Improvement of Image Generation Technology}
Early models suffered from low image quality and poor adherence to text prompts, making compositional control difficult. Recent models demonstrate improved text understanding by incorporating multiple text encoders. For compositional control, the introduction of ControlNet \cite{zhang2023adding} in 2023 enabled image generation that preserves contours, poses, and depth information from a source image, significantly expanding the controllability of the generation process. This marked an evolution from simple ``drawing AIs" to instructable tools for design and media production.

Furthermore, the adoption of efficient fine-tuning methods like LoRA (Low-Rank Adaptation) \cite{hu2021lora}, which allows for additional training on specific characters or styles without retraining the entire model, has enabled individuals to create customized models, spanning both general-purpose and specialized applications.

Entering 2024, architectures such as the Diffusion Transformer (DiT) \cite{peebles2022scalable}, which integrates the Transformer architecture that proved successful for LLMs into diffusion models, are becoming prevalent. Stable Diffusion 3 \cite{stability2024sd3, stability2024sd3paper} follows this trend, offering markedly improved text interpretation (prompt fidelity) over conventional U-Net-based models and enabling more complex and accurate image generation. These technologies are detailed in Section \ref{sec:competing}.

\subsection{Ethical Issues and Dataset Constraints in Image Generation AI}
Early models, including Stable Diffusion, were trained on large-scale datasets (e.g., LAION \cite{schuhmann2022laion5b}) containing billions of image-text pairs scraped from the public web, without obtaining individual licenses from rights holders. The nature of such training data has sparked legal and ethical debates. In several countries, rights holders have filed lawsuits over models trained on datasets that may contain materials with unconfirmed rights status or illegally uploaded content, highlighting the legal risks involved.

Legal frameworks remain inconsistent globally. In Japan, Article 30-4 of the Copyright Act \cite{egov2018copyright, bunka2024ai} permits the use of copyrighted works for information analysis under specific conditions, without individual permission from the rights holder. However, ambiguities in legal interpretation and differing views on copyright persist, and case law is insufficient. In practice, it is desirable to select datasets with the lowest possible rights-related risks. While T2I development was initially driven by the OSS community for research and personal use, its adoption in business is expanding. This has led to friction, with some companies and public organizations facing public criticism over their use of generative AI.

This research focuses on developing a model for anime production. The use of AI trained on unlicensed data in animation may not be viewed positively by audiences. Furthermore, creators face the challenge of avoiding models that carry a non-trivial risk of coincidentally generating outputs similar to existing copyrighted works. However, curating large-scale datasets with clear, verifiable rights is difficult. This necessitates a model architecture capable of effective learning from limited data.

\subsection{Performance and Training Dataset Requirements for Image Generation AI Introduced in Anime Production}
The model developed in this project is intended for adoption by anime production companies. As performance requirements and data confidentiality needs differ between companies, customized models are necessary. Our development strategy assumes a two-stage process: first, training a ``foundation model" on a dataset with verified rights, and second, applying ``specialized model" training for each company using their proprietary data.

In this document, a ``dataset with verified rights" (or ``copyright-cleared dataset") refers to one composed of images in the Public Domain, under CC0, or used with explicit permission for training from the copyright holders. Verifying each image reduces the available data volume compared to datasets like LAION. This scarcity can lead to overfitting or poor generation quality, necessitating a new model architecture optimized for smaller datasets.

\section{Status of Competing Models}
\label{sec:competing}
The T2I generation field was pioneered by Generative Adversarial Networks (GANs) \cite{goodfellow2014gan}. GANs utilize adversarial learning, pitting a ``Generator" against a ``Discriminator." This architecture can generate images rapidly post-training but suffers from training instability, ``Mode Collapse" (a loss of output diversity), and vanishing gradients.

Diffusion Models \cite{ho2020ddpm} emerged to address these challenges. Inspired by non-equilibrium thermodynamics, they learn a ``forward process" (adding noise) and a ``reverse process" (denoising). This iterative denoising ensures stable training and superior sample diversity compared to GANs, though at the cost of higher inference computation.

Modern T2I models (e.g., Stable Diffusion, Midjourney, DALL-E \cite{ramesh2021zero}) are based on diffusion architectures. They typically comprise three components:
\begin{enumerate}
    \item \textbf{Text Encoder}: Converts the input text prompt into embeddings.
    \item \textbf{Image Inference Architecture}: Executes the denoising process in latent space based on text conditions.
    \item \textbf{Variational Autoencoder (VAE)}: Compresses images into a low-dimensional latent space and reconstructs them back to pixel space.
\end{enumerate}

\subsection{Text Encoders: Deepening Semantic Understanding}
Text encoders vectorize input prompts for the neural network. Performance depends on capturing the content, order, and relationships within the text, which is crucial for prompt fidelity. The output vector is used for ``conditioning" the generation process.

The original Stable Diffusion 1.5 (SD1.5) adopted CLIP \cite{radford2021learning}, but its lightweight nature limited performance. Newer models often supplement CLIP with larger encoders like T5 or Gemma.

\subsubsection{CLIP: Achieving Conceptual Alignment}
CLIP (Contrastive Language-Image Pre-training) \cite{radford2021learning} learns a shared embedding space for images and text through large-scale contrastive learning on 400 million image-text pairs. It maps semantically similar text and images to nearby points in this space (e.g., the text ``a photo of a dog" and an actual dog image). This allows CLIP to perform ``zero-shot learning." In T2I models, CLIP links textual concepts (e.g., ``astronaut") and styles (e.g., ``photo") to visual features. Models like SD1.5 and Stable Diffusion XL (SDXL) \cite{podell2023sdxl} use CLIP as a core text encoder.

\subsubsection{T5: Precise Interpretation of Language Structure}
CLIP has limitations in understanding complex syntax and rendering text (typography). T5 (Text-to-Text Transfer Transformer) \cite{raffel2019exploring}, developed by Google, gained attention as a solution to this limitation.

T5 is an Encoder-Decoder Transformer that handles all NLP tasks as text-to-text problems. Pre-trained on the C4 corpus \cite{c4dataset}, it possesses a deep understanding of grammar, syntax, and context. In T2I models, T5's advantage is its advanced language comprehension. It can more accurately interpret spatial relationships (e.g., ``a red cube on top of a blue sphere") or specific text (e.g., ``graffiti reading `Stable Diffusion'") than CLIP. Consequently, recent models like Stable Diffusion 3 and FLUX.1 typically use T5 in combination with CLIP. While decoder-only LLMs have recently been adapted as text encoders (see Section \ref{sec:gemma}), the value of encoder-decoder models is being reconsidered \cite{zhang2025encoder}. We adopted the T5 v1.1 XXL encoder-decoder based on favorable experimental results.

\subsubsection{Gemma and Other Encoders}
\label{sec:gemma}
The open-sourcing of LLMs has expanded encoder options. Gemma \cite{gemma2024}, a lightweight open model family from Google, can process multimodal inputs and is being adopted in new T2I models. Lumina-Image 2.0 \cite{qin2025lumina}, for example, uses Gemma-2-2B.

This trend indicates a shift away from a single encoder toward a ``committee of experts" approach. SD1.5 (CLIP-only) excelled at concepts but failed at complex compositions. SDXL (two CLIPs) improved nuance but not structural understanding. A significant breakthrough came with the hybrid (CLIP + T5) approach used by Stable Diffusion 3 and FLUX.1. In this setup, CLIP defines ``what" to draw (concepts, style), while T5 deciphers ``how" to draw it (syntax, spatial relations, spelling). This division of roles allows the encoders to compensate for each other's weaknesses, achieving high prompt fidelity and typography performance.

\subsection{Image Inference Architecture: From U-Net to Transformer}
The image inference architecture converts the text embeddings into pixel information. This component executes the diffusion denoising process, and its design determines the model's performance, scalability, and quality. The main trend has been a transition from the CNN-based U-Net to the more scalable Transformer.

\subsubsection{U-Net: The Foundation of Convolutional Networks}
U-Net \cite{ronneberger2015unet}, originally for medical image segmentation, features a ``U-shaped" encoder-decoder structure. The encoder uses convolutions and pooling to extract features while downsampling, and the decoder reconstructs the image via transposed convolutions. Its key feature is the ``skip connection," which passes high-resolution positional information from the encoder directly to the corresponding decoder layer, enabling precise reconstruction.

In diffusion models, the U-Net predicts and removes noise at each timestep. Mainstream models up to circa 2023 (e.g., SD1.5, SDXL) used modified U-Nets \cite{ho2020ddpm} incorporating mechanisms like self-attention.

\subsubsection{DiT (Diffusion Transformer): Achieving Scalability}
The inductive biases of U-Nets are not strictly necessary for diffusion models. Research \cite{peebles2022scalable} demonstrated that Transformers, leveraging their proven success in NLP and vision, could serve as a scalable backbone for diffusion models. This architecture, the Diffusion Transformer (DiT), has since become common. While the original DiT used class-label conditioning, adaptations for text-conditioning (e.g., cross-attention, joint attention) are widely used.

In a DiT, the VAE-compressed latent image is divided into patches, which are fed to the Transformer as a token sequence. This allows the model to learn both local (intra-patch) and global (inter-patch) features.

The most important characteristic of DiT is its scalability. Experiments show that image quality (measured by FID) scales predictably with model size and computation. This scaling property underpins next-generation models like OpenAI's Sora and Stable Diffusion 3.

\subsection{VAE: Representation and Reconstruction in Latent Space}
In Latent Diffusion Models (LDMs) like the Stable Diffusion family, the VAE is essential for computational efficiency and final image quality.

\subsubsection{The Role of VAE in Latent Diffusion Models}
Directly processing high-resolution images (e.g., 1024x1024) in pixel space is computationally prohibitive. LDMs solve this by using a bottleneck VAE to compress the image into a lower-dimensional ``latent space," where the diffusion and denoising occur.
\begin{enumerate}
    \item \textbf{Encoding}: The VAE encoder converts a high-resolution image (e.g., 512x512x3) into a low-dimensional latent representation (e.g., 64x64x4).
    \item \textbf{Decoding}: After the inference architecture (U-Net or DiT) completes denoising, the VAE decoder reconstructs the clean latent representation back into a high-resolution pixel image.
\end{enumerate}
This latent-space operation significantly reduces the computational load and memory usage.

\subsubsection{The Trade-off between Compression Rate and Detail Preservation}
The VAE's design, particularly its ``compression rate," is a critical bottleneck for final image quality. VAE compression is lossy; some information is irreversibly lost.

Common artifacts like blurry faces, unnatural hands, or illegible text are often attributed to the denoiser, but they may originate from the VAE compression stage. If fine detail is lost during encoding, no subsequent denoiser can recover it.

This problem is tied to the VAE's channel count. The standard VAE used in SD1.5 and SDXL compresses to a 4-channel latent space. The SDXL VAE, in particular, has a high 1:48 compression rate, prioritizing efficiency over fidelity.

To resolve this bottleneck, recent models like FLUX and Stable Diffusion 3 utilize a 16-channel VAE. The FLUX VAE, for instance, has a 1:12 compression rate. By lowering the compression, the VAE retains more fine-grained information (e.g., facial expressions, textures, small text) in the latent space. Evaluations of 16-channel VAEs show quantitative improvements in reconstruction metrics (PSNR, LPIPS) over 4-channel VAEs \cite{auradiffusion2024vae}.

Thus, VAE evolution is a fundamental improvement that raises the quality ceiling of the entire pipeline, reflecting a paradigm shift from prioritizing speed to prioritizing fidelity.

\subsection{Stable Diffusion 1.5}
Stable Diffusion 1.5 (SD1.5) \cite{rombach2022highres}, released by RunwayML in 2022, is a foundational model that popularized open-source, high-performance T2I generation.
\begin{itemize}
    \item \textbf{Text Encoder}: CLIP ViT-L/14 \cite{clipvitlarge}. A single CLIP encoder converts English prompts into 768-dimensional embeddings for conditioning.
    \item \textbf{Image Inference Architecture}: U-Net. A standard U-Net \cite{peebles2022scalable} for latent diffusion, using cross-attention to integrate text conditions.
    \item \textbf{VAE}: Standard 840k-step VAE. Compresses 512x512 images to a 64x64x4 latent space. This VAE was known for producing slightly desaturated and blurred details. (Later, the OSS community trained improved VAEs).
\end{itemize}
Trained on a subset of LAION, SD1.5 established the LDM ``trinity" (CLIP, U-Net, VAE) and provided (txt2img, img2img, inpainting) under the OpenRAIL-M license, fostering a vast ecosystem.

\subsection{Stable Diffusion XL (SDXL) and Refiner}
Stable Diffusion XL (SDXL) \cite{podell2023sdxl} is the successor to SD1.5, designed for higher resolution and better prompt adherence, introducing an ``ensemble of experts" concept.
\begin{itemize}
    \item \textbf{Text Encoder}: Uses two: CLIP ViT-L/14 and the larger OpenCLIP ViT-bigG/14 \cite{openclipvitbigG}. The concatenated outputs provide richer text representations.
    \item \textbf{Image Inference Architecture}: A larger U-Net (approx. 3x parameters of SD1.5) operating natively at 1024x1024 resolution.
    \item \textbf{VAE}: SDXL VAE (4-channel). An improved VAE for 1024x1024 images, though some users reported artifacts and preferred community-tuned alternatives.
\end{itemize}
SDXL's dataset is not disclosed but is presumed to include LAION and ImageNet. It proposed a two-stage pipeline:
\begin{itemize}
    \item \textbf{Base Model}: Generates the primary 1024x1024 latent representation.
    \item \textbf{Refiner Model}: A separate, smaller LDM that applies further denoising to the base model's output to add high-frequency details. The Refiner's improvements were considered limited, and it is not widely used today.
\end{itemize}

\subsection{Stable Diffusion 3 (SD3)}
Stable Diffusion 3 (SD3) \cite{stability2024sd3, stability2024sd3paper} is Stability AI's flagship model, featuring a revamped architecture.
\begin{itemize}
    \item \textbf{Text Encoder}: Uses three: CLIP ViT-L, OpenCLIP ViT-G, and the 4.7B-parameter T5-XXL. This combination balances visual concepts (CLIPs) with linguistic structure (T5).
    \item \textbf{Image Inference Architecture}: MMDiT (Multimodal Diffusion Transformer). Based on DiT, MMDiT uses separate Transformer weights for image and text modalities, which are combined only during the attention calculation. This bidirectional flow improves text-image alignment, especially for typography. It also adopts Rectified Flow for faster inference.
    \item \textbf{VAE}: New 16-channel VAE. Adopted to minimize information loss and faithfully render details like hands and faces.
    \item \textbf{Dataset}: Not disclosed.
\end{itemize}
\textit{Features}: SD3 integrates MMDiT, a three-encoder setup, Rectified Flow, and a 16-channel VAE. It offers multiple sizes (800M to 8B parameters).

\subsection{Stable Diffusion 3.5 (SD3.5)}
Stable Diffusion 3.5 (SD3.5) \cite{stability2024sd35} is the latest refinement of the SD3 architecture, focusing on training stability and performance.
\begin{itemize}
    \item \textbf{Text Encoder}: Same as SD3 (CLIP-L, CLIP-G, T5-XXL).
    \item \textbf{Image Inference Architecture}: MMDiT-X. An improved MMDiT (``X" for improvements) incorporating QK-normalization \cite{wang2023qknorm} for training stability and dual attention blocks in the initial 12 layers.
    \item \textbf{VAE}: Same 16-channel VAE as SD3.
    \item \textbf{Dataset}: Not disclosed.
\end{itemize}
\textit{Features}: SD3.5 is a solid update, employing progressive training (256x256 up to 1440x1440) and mixed-resolution training to improve aspect ratio flexibility.

\subsection{FLUX.1}
FLUX.1 \cite{bfl2025flux, bfl2024announcing}, from Black Forest Labs, is a next-generation T2I model prioritizing speed and fidelity.
\begin{itemize}
    \item \textbf{Text Encoder}: Uses CLIP and T5. Similar to SD3, CLIP handles visual concepts while T5 handles linguistic structure and spelling.
    \item \textbf{Image Inference Architecture}: Rectified Flow Transformer (12B parameters). FLUX.1 uses Rectified Flow, based on Ordinary Differential Equations (ODEs) rather than Stochastic Differential Equations (SDEs). This linearizes the path from noise to data, drastically reducing required inference steps (1-4 steps for the 'Schnell' version).
    \item \textbf{VAE}: FLUX VAE (16-channel). A 16-channel VAE with low compression (1:12) to retain fine details like text and textures.
    \item \textbf{Dataset}: Not disclosed.
\end{itemize}
\textit{Features}: FLUX.1 combines Rectified Flow (speed), a CLIP+T5 hybrid (fidelity), and a 16-channel VAE (detail). The 'Kontext' model variant also allows image+text inputs for editing.

\subsection{Lumina-Image 2.0}
Lumina-Image 2.0 \cite{qin2025lumina}, by Alpha-VLLM, is an advanced framework focused on architectural ``unification" and ``efficiency."
\begin{itemize}
    \item \textbf{Text Encoder}: Gemma-2-2B, Google's open-source LLM.
    \item \textbf{Image Inference Architecture}: Unified Next-DiT. This architecture abandons cross-attention. Instead, it concatenates text tokens and image (latent) tokens into a single sequence. Joint Self-Attention is applied to this unified sequence, allowing text and image information to interact bidirectionally, analogous to a decoder-only LLM.
    \item \textbf{VAE}: FLUX-VAE-16CH. Uses the high-fidelity 16-channel FLUX VAE.
    \item \textbf{Dataset}: Not disclosed.
\end{itemize}
\textit{Features}: The ``unification" philosophy treats text and images as a single information stream, avoiding the biases of cross-attention. This design is inherently scalable to other modalities (e.g., audio, video).

\subsection{Issues with Existing Models and Required Specifications for a New Model}
As of September 2025, SDXL is widely used, largely due to its licensing and hardware requirements. Most newer models have restrictive licenses (e.g., non-commercial) or are too large for consumer hardware, hindering adoption.

For the anime industry, dataset provenance is a major barrier. SD1.5 used the LAION dataset, which has known ethical concerns. Most other models do not disclose their training data. Using such models in commercial anime production poses significant risks and is a barrier to adoption.

A new model must address these issues: it must use a disclosed, copyright-cleared dataset. However, such datasets are far smaller than LAION (millions vs. billions of images). This data scarcity necessitates a model architecture that can achieve high performance from a small dataset.

Furthermore, rights issues may be discovered in a dataset post-training. Removing specific data from a trained model is difficult, often requiring retraining from scratch. Therefore, a foundation model that learns efficiently from less data is required.

\section{Developed Model}
\subsection{Overall Model Configuration}
oboro: is a text-conditioned image generation diffusion model. It uses the I-CFM flow formulation and a custom DiT architecture featuring a Multi-Multi-Head Attention mechanism (Section \ref{sec:network-details}). It employs the T5 V1.1 XXL text encoder and the FLUX VAE.

\subsection{Details of Network Structure}
\label{sec:network-details}
\begin{figure}[htbp]
  \centering
  \includegraphics[width=8cm]{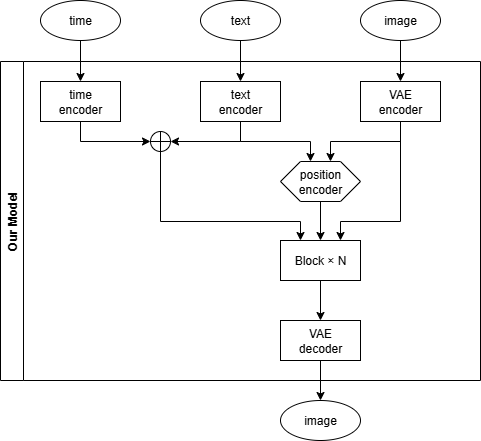}
  \caption{Our Model: Overall Structure. Inputs (time, text, image) are processed by encoders, combined, passed through N Blocks, and decoded by the VAE decoder.}
  \label{fig:model_structure}
\end{figure}

\begin{figure}[htbp]
  \centering
  \includegraphics[width=6cm]{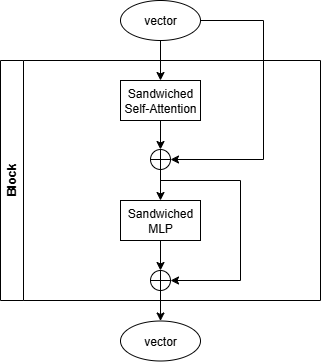}
  \caption{Block Structure, showing Sandwiched Self-Attention and Sandwiched MLP with residual connections.}
  \label{fig:block_structure}
\end{figure}

\begin{figure}[htbp]
  \centering
  \includegraphics[width=6cm]{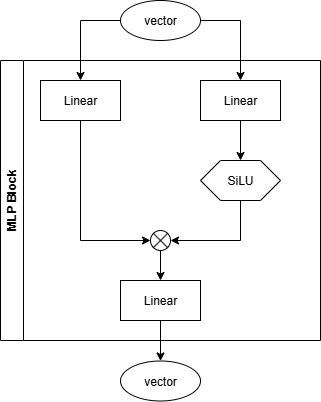}
  \caption{MLP Block Structure, showing Linear layers, SiLU activation, and element-wise multiplication.}
  \label{fig:mlp_block}
\end{figure}

\begin{figure}[htbp]
  \centering
  \includegraphics[width=8cm]{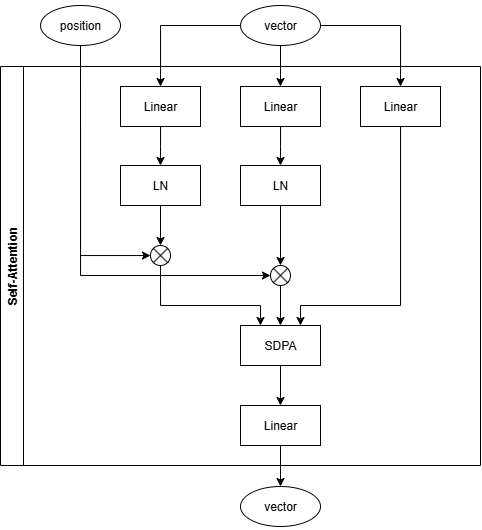}
  \caption{Self-Attention Structure, showing inputs (position, vector), Linear layers, Layer Norm (LN), SDPA, and position modulation.}
  \label{fig:self_attention}
\end{figure}

\begin{figure}[htbp]
  \centering
  \includegraphics[width=8cm]{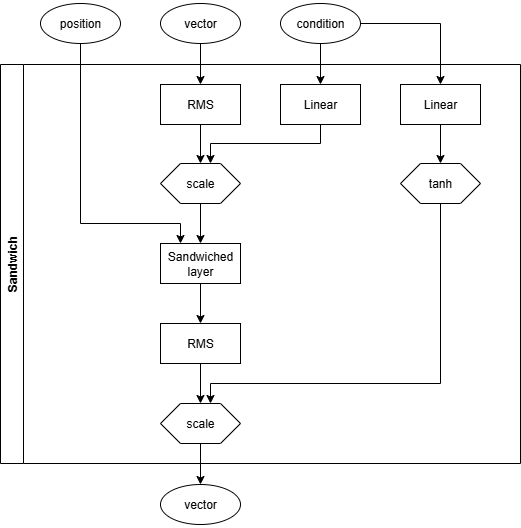}
  \caption{Sandwich Normalization Structure, showing RMS, scaling, and conditioning.}
  \label{fig:sandwich}
\end{figure}

\begin{figure}[htbp]
  \centering
  \includegraphics[width=5cm]{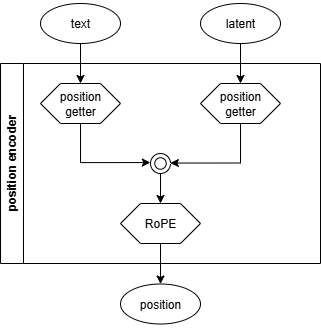}
  \caption{Position Encoder Structure, showing position getters for text and latent, concatenation, and RoPE.}
  \label{fig:pos_encoder}
\end{figure}

The model's network structure is illustrated in Fig. \ref{fig:model_structure} through \ref{fig:pos_encoder} (placeholders).

\subsubsection{Diffusion Transformer (DiT)}
We adopted a DiT structure similar to those in SD3, FLUX.1, Lumina-Next-T2I \cite{zhuo2024lumina}, Lumina-Image 2.0, and Mochi \cite{mochi2024}. The encoder maps time, input text, BOS token, input image, and position IDs to embeddings via 3-layer MLPs or fully-connected layers. The Transformer backbone employs SwiGLU nonlinearity \cite{shazeer2020glu}, 2D-RoPE \cite{heo2024rope, su2021roformer}, Sandwich normalization \cite{shleifer2021normformer, ding2021cogview}, QK normalization \cite{wang2023qknorm}, and Joint Self-Attention \cite{qin2025lumina}.

\subsubsection{Multi-Multi-Head Attention}
We introduced a specific design termed Multi-Multi-Head Attention into our multi-layer DiT architecture. This method assigns a different number of attention heads to the attention layer within each DiT block.

This design is inspired by the hierarchical feature processing of U-Net. In a standard stack of identical DiT blocks, role differentiation among blocks may not occur efficiently, potentially slowing feature extraction in early training phases. For example, if all blocks attempt to learn both fine details and global structure simultaneously, convergence may be hindered. U-Net's architecture inherently manages this separation of concerns.

To address this, Multi-Multi-Head Attention encourages an implicit division of roles by varying the head count, thereby controlling the information granularity \cite{vaswani2017attention}. We determined the optimal head count allocation through validation experiments with fewer layers. An improvement in learning efficiency was confirmed when progressively increasing the layer count using this scheme.
\begin{itemize}
    \item \textbf{Early DiT blocks (few heads)}: Set to 8 or 16 heads. A lower head count assigns a wider subspace to each head, encouraging the model to capture global, low-frequency features (e.g., layout, object relations).
    \item \textbf{Later DiT blocks (many heads)}: Set to 24 or 48 heads. More heads allow for specialization in narrower subspaces, assigning these layers the role of refining local, high-frequency details (e.g., textures, contours).
\end{itemize}
The 8/16/24/48 head configuration performed well in experiments and was adopted for the final training.

\subsection{Learning Strategy and Loss Function}
We adopted I-CFM (Independent Conditional Flow Matching) as the learning strategy and MSE (mean squared error) as the loss function. This method is similar to Rectified Flow \cite{liu2022flow}, which learns the flow along a linear interpolation path $x_t = tx_1 + (1-t)x_0$ between a data sample $x_1$ and a noise sample $x_0$.

OTCFM \cite{tong2023improving}, which uses Optimal Transport (OT) to streamline flow learning, was proposed to reduce discretization error and enable few-step inference. However, in our small-scale experiments combined with text conditioning, it did not yield sufficient performance. We therefore adopted I-CFM, also proposed in the OT-CFM paper. I-CFM is equivalent to adding a noise term to the Rectified Flow, making inference more robust, and has been used in models like Macha-TTS \cite{mehta2023matchatts}.

\subsection{Selection of Text Encoder}
Candidates for the text encoder included Gemma-2 2B, Gemma-2 2B (rinna-FT), Gemma-2 9B, Quen-2 1.5B, Quen-2 3B, Quen-2-VL 2B, Phi-3.5-mini-inst, T5-1.1-XXL, CLIP-ViT L/14, SIGLIP \cite{zhai2023siglip}, and SIGLIP-so400M. We ultimately adopted T5 Version 1.1 XXL \cite{raffel2019exploring, t5v11xxl}, as small-scale experiments demonstrated it generated images that better reflected the input text.

\section{Learning Process}
\subsection{Dataset}
\subsubsection{Selection}
This project targets anime production assistance, requiring a model that user companies can deploy with confidence. This mandates the use of a copyright-cleared dataset. As noted in Section \ref{sec:competing}, most competing models use unlicensed data or do not disclose their data sources. To ensure the model is commercially viable for anime businesses, we selected a dataset with verified rights.

From this perspective, we selected the Megalith-10m dataset \cite{megalith10m}. It was chosen because its construction process exclusively used data sources with clear copyright status (CC0-equivalent images from Flickr), providing a relatively large-scale and diverse image collection.

\subsubsection{Deduplication}
The Megalith-10m dataset contained many duplicates. To prevent overfitting, training instability, and output bias, we performed duplicate removal.

We implemented an iterative clustering method using CLIP image embeddings. First, a pre-trained CLIP model (DataComp L/14) was used to extract embeddings for all images.

To reduce computational load, the embeddings were partitioned into subsets. DBSCAN clustering (cosine similarity, eps=0.9) was run on each subset. The threshold was empirically set to group similar images.

After clustering, a set of representative images for each subset was created by combining one randomly selected image from each cluster and all noise images (those not assigned to any cluster).

These representative images were integrated into a new set, which was then re-partitioned, and the process was repeated. This iterative cycle continued until the set of representative images converged, ensuring comprehensive deduplication.

\begin{figure}[htbp]
  \centering
  \includegraphics[width=8cm]{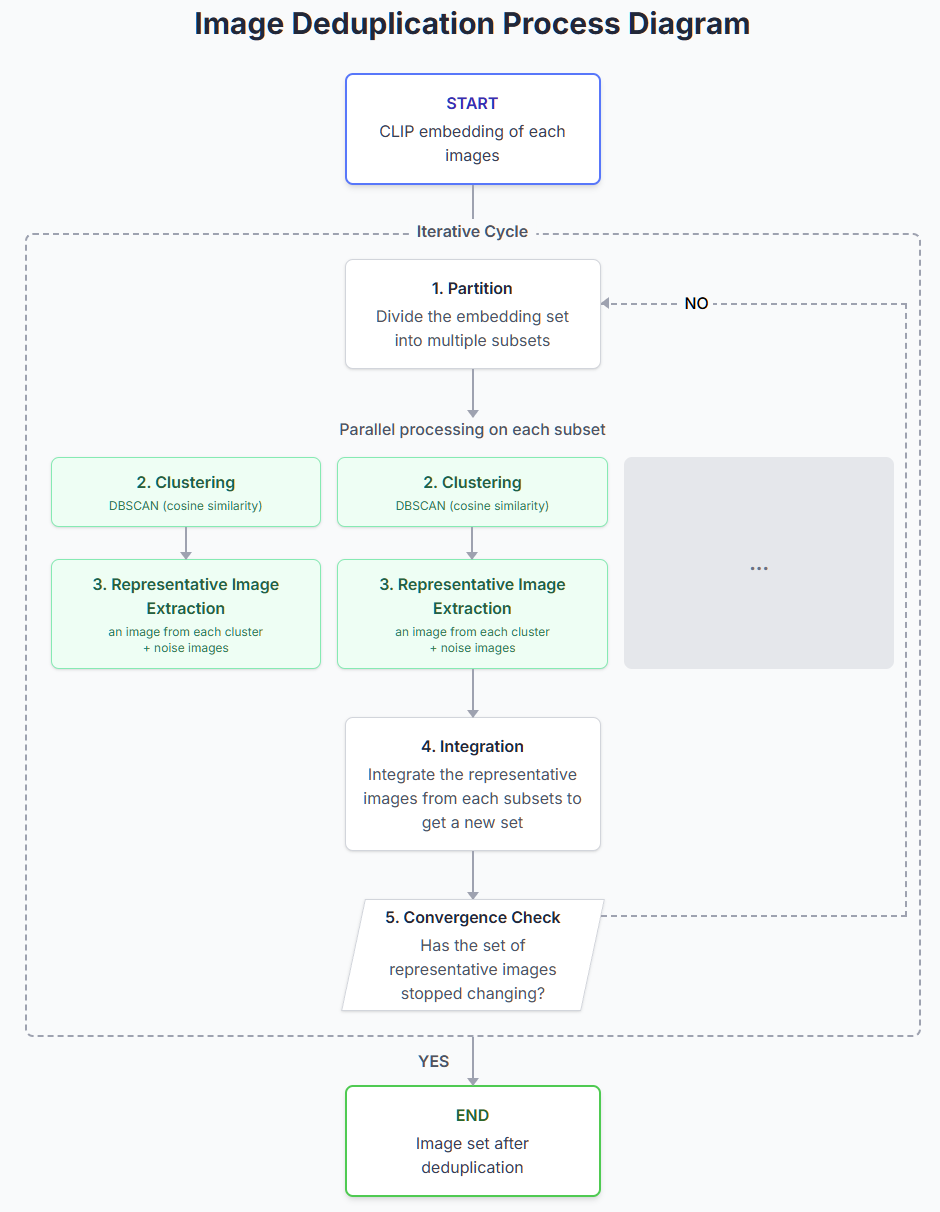}
  \caption{Image Deduplication Process Diagram. An iterative cycle of partitioning, DBSCAN clustering, representative extraction, integration, and convergence check.}
  \label{fig:deduplication}
\end{figure}

\subsubsection{Captioning}
Megalith-10m is an image-only dataset, though third-party captions like ShareCaptioner \cite{sharecaptioner} exist. We generated custom captions to include multiple tag types.

We compared several VLMs (Florence-2, Molmo, WD 1.5 Tagger, Pixtral, Phi-3.5 vision instruct, Qwen) and selected Florence-2 \cite{florence2}. It was chosen for its superior understanding of anime-style expressions, considerably faster generation speed, and its ability to reliably generate captions of varying lengths, which suited our objectives. Speed was a priority for captioning $\sim$10 million images. WD 1.5 Tagger, while fast, produced tag lists unsuitable for our intended use.

\subsubsection{Scoring}
We added a dedicated tag to each caption indicating image quality. This involves categorizing dataset images by quality and adding specific tags, allowing users to improve output by adding a ``high quality" tag to the prompt (or a ``low quality" tag to the negative prompt).

We used Aesthetic Predictor V2.5 \cite{aestheticpredictor}, a SigLIP-based tool respected by model creators, to score images on a 1-10 scale. Tags were added based on scores: 10-6 (``excellent score," top $\sim$3\%), 6-5.2 (``good score," top $\sim$20\%), and 5.2-4 (``average score," top $\sim$80\%). Images scoring below 4.0 (approx. 20\% of the data) were excluded based on visual inspection of random samples.

\subsubsection{Conversion to Latent}
To save computational resources during training, all images were pre-converted into latent representations using the VAE encoder. This eliminated redundant VAE encoding computations during the training loop and reduced the data transfer bandwidth required from main memory to VRAM. This pre-processing was performed on local machines in parallel with early training tests, improving overall training efficiency.

\subsubsection{Two-Stage Resolution Setting}
Following common practice \cite{rombach2022highres, peebles2022scalable, stability2024sd3, ramesh2021zero, podell2023sdxl, zhuo2024lumina, stability2024sd3paper}, we used multi-stage resolution training. Images were processed in two stages: (1) resizing the short edge to 256 pixels and applying a random crop to 256x256, and (2) resizing to match an area of $\sim$250k pixels (e.g., 512x512) with an aspect ratio < 1:3. Images were then cropped again to ensure dimensions were multiples of 64. All resizing maintained the aspect ratio (excluding sub-pixel discretization errors).

\subsection{Learning Environment}
\subsubsection{Hardware}
We used an AWS ParallelCluster environment with H100x8 instances. We observed stability issues within this environment, managed by the Slurm job scheduler. GPU processes would halt at irregular intervals (hours to days). Reproduction attempts suggested the cause was not isolated to the training code. Despite vendor technical support, a permanent solution was not found during this phase. Similar issues with this hardware configuration have been reported externally \cite{pfntechblog2024}. This instability impacted throughput and scheduling, requiring 24/7 monitoring and recovery efforts to maintain $\sim$80\% effective efficiency. This increased the operational load and posed a project management risk. This experience suggests that future projects using similar configurations must prioritize infrastructure stability, pre-deployment verification, and redundancy.

\subsubsection{Software}
\paragraph{Work Environment Construction}
We used Rye \cite{rye2025}, a unified Python toolchain manager, and uv \cite{uv2025}, a high-speed package installer, to build the development environment.

\textit{Rye}: Rye \cite{rye2025} consolidates Python version management, virtual environments, and dependency management into a single CLI, simplifying the setup previously handled by separate tools (pyenv, venv, pip).

\textit{uv}: uv \cite{uv2025} is a fast Python package installer and resolver implemented in Rust. As Rye's default installer, it significantly shortens dependency resolution and installation times. This speed was highly beneficial for rapid iteration.

\textit{Effects of Introduction}: The use of Rye and uv improved setup efficiency. Team members could use a single \texttt{rye} command for environment setup, reducing time and cognitive load. The speed of uv accelerated development cycles, especially when managing large libraries. These tools contributed to a smooth project start and consistent environments across developers.

\paragraph{PyTorch Lightning}
We initially adopted PyTorch Lightning to abstract boilerplate code and accelerate research. However, we encountered challenges as development progressed. Customizing data aggregation and shared processes derived from PyTorch Distributed proved difficult. The abstraction layer sometimes became a constraint when fine-grained control over distributed learning was required. Debugging was often difficult due to the lack of visibility into Lightning's internal processes. Based on this experience, using PyTorch Distributed directly might have been more
suitable for our requirements.

\paragraph{Weights \& Biases (W\&B)}
We used Weights \& Biases (W\&B) for experiment tracking and monitoring. This allowed for efficient visualization of metrics (loss, accuracy), hyperparameter tracking, and generated image inspection, improving reproducibility and analysis.

\subsection{Hyperparameters}
\subsubsection{Architecture}
The number of Transformer layers was set to 32. Performance scaled positively up to 32 layers; we did not increase it further to conserve VRAM and ensure model manageability.
\begin{itemize}
    \item \textbf{Non-Trainable Layers}: The pre-trained T5 V1.1 XXL text encoder, the pre-trained FLUX VAE, and the RoPE encoder (which requires no training).
    \item \textbf{Trainable Layers}: Non-frozen layers in the encoder block, the 32 intermediate Transformer layers, and the final Transformer layer.
    \item \textbf{Patch Size}: The pixel unshuffle \cite{shi2016realtime} factor (patch size) was set to 2, consistent with SD3 and FLUX.1 \cite{stability2024sd3, bfl2025flux}, as a balance for computational load.
\end{itemize}

\subsubsection{Optimizer}
We used the AdamW optimizer \cite{loshchilov2017decoupled} with a warmup scheduler throughout training. This combination is standard for large-model training and offers proven stability. We prioritized stability over exploring novel optimizers.

Memory-efficient optimizers (e.g., Adafactor, Lion) were considered, but VRAM was not a bottleneck in our environment, so their benefits were limited. ScheduleFree \cite{defazio2024road} was tested but showed no clear advantage over AdamW with warmup.

The AdamW parameters $\beta_1$, $\beta_2$, and $\epsilon$ were set to 0.9, 0.999, and 1e-8, respectively.

\subsubsection{Learning Rate}
The learning rate for the 256x256 pre-training stage (Section \ref{sec:staged-learning}) was 2e-4. For the 512x512-area stage, the learning rate started at 1e-4 and was decreased to 7e-5. For the final additional learning stage, it started at 2e-5 and was gradually decreased to 1e-6. These values were determined empirically by monitoring training.

\subsubsection{Batch Size}
In the 256x256 stage, the global batch size was 1024 (2 nodes $\times$ 8 GPUs/node $\times$ 64 samples/GPU). In the 512x512-area stage, the global batch size was 384 (2 nodes $\times$ 8 GPUs/node $\times$ 24 samples/GPU).

\subsection{Learning Execution}
\subsubsection{Small-Scale Learning Tests}
The DiT architecture allows model size adjustment by changing the number of layers. Prior to full-scale training, we conducted preliminary tests using a lightweight model (20 Transformer blocks) to validate the training pipeline and perform initial hyperparameter exploration efficiently.

\subsubsection{Staged Learning}
\label{sec:staged-learning}
Pre-training was conducted in two stages: (1) 256x256 square images and (2) 512x512-area rectangular images ($\sim$250k pixels). Additional specialized training was also performed using 512x512-area images.

\subsubsection{Learning Convergence Process}
\begin{figure}[htbp]
  \centering
  \includegraphics[width=8cm]{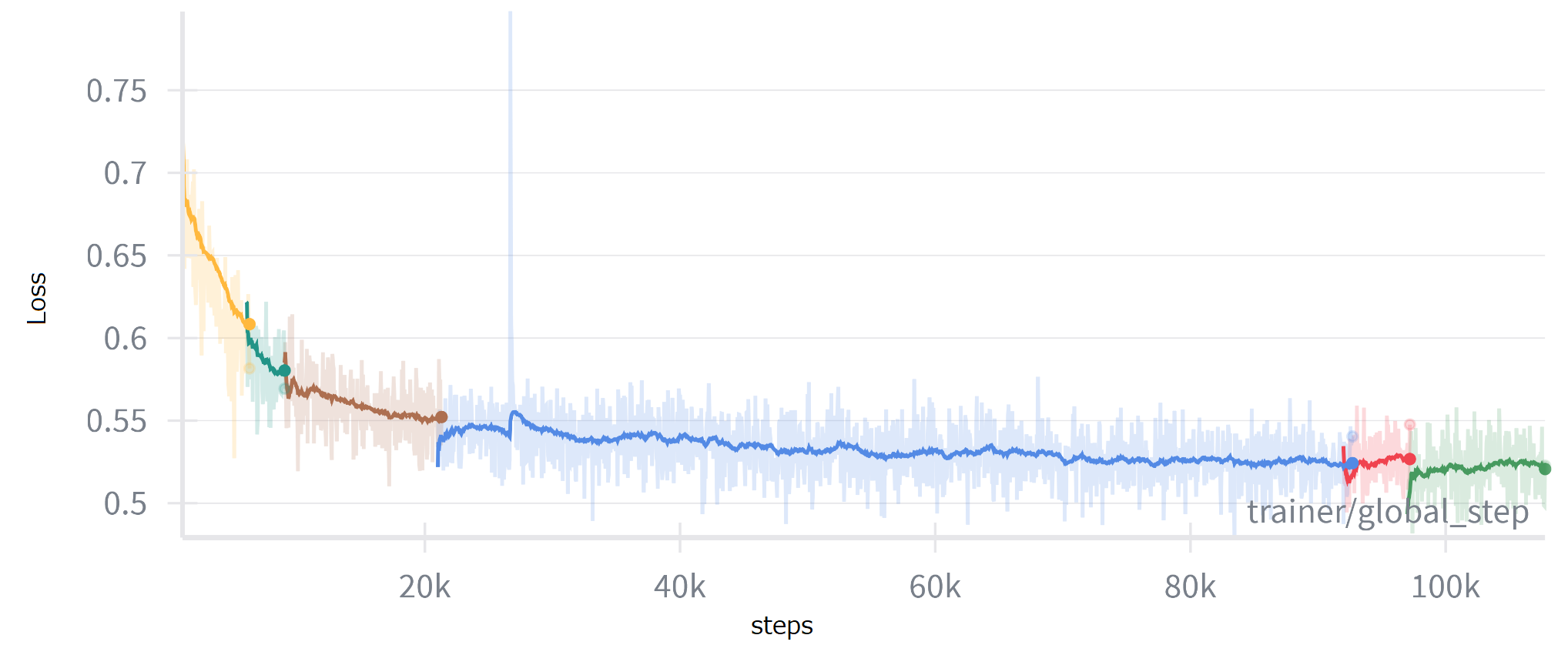}
  \caption{Training loss curve (train/loss) for 256x256 pre-training, as recorded by W\&B. The graph shows several training runs (e.g., `royal-mountain-225`) and interruptions due to hardware issues.}
  \label{fig:loss_curve}
\end{figure}

Fig. \ref{fig:loss_curve} shows the 256x256 pre-training loss, logged by W\&B (light line: raw data; dark line: EMA, 0.9).

A loss spike early in the `royal-mountain-225` run was visually inspected; as generated images showed no issues, training was continued. The interruptions visible in the graph correspond to the hardware issues described in Section IV-B1.

\subsubsection{Measures Taken to Stabilize Learning}
While the 256x256 pre-training was stable, the 512x512-area intermediate training and the anime-specialized training stages exhibited frequent loss spikes. We addressed this instability by manually applying a gradual learning rate decay. Standard schedulers like Cosine decay failed to stabilize training. In our exam, ZClip\cite{kumar2025zclip} have not worked to suppress loss spikes. Future work may explore optimization methods like ScheduleFree \cite{defazio2024road}.

\subsection{Communication and Information Sharing}
Project communication was centralized on Discord for real-time discussion and information flow. Regular meetings, ad-hoc discussions, and paper sharing occurred on the platform. Stock knowledge (e.g., research summaries) was managed in Notion, while official reports were co-edited in Google Docs. We integrated Discord with Hugging Face and W\&B repositories and deployed a bot for automated notifications of training progress and errors.

\section{Experiments and Evaluation}
\subsection{Qualitative Evaluation of Generation Results}
The developed foundation model is named ``oboro:base." The ``oboro:" prefix is common, with specialized models intended to be named ``oboro:[CompanyName]."

\begin{figure}[htbp]
  \centering
  \includegraphics[width=8cm]{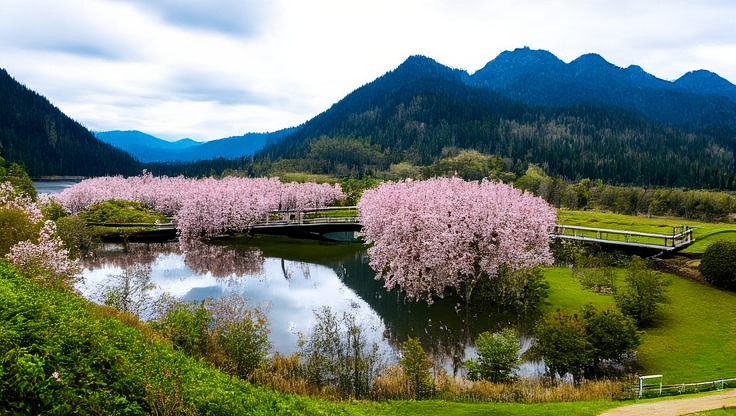}
  \caption{Example output of ``oboro:base".}
  \label{fig:example_output}
\end{figure}

Fig. \ref{fig:example_output} shows an output from ``oboro:base." As this model was trained exclusively on photorealistic images, it generates realistic scenes. The prompt used is:
\begin{verbatim}
A serene spring landscape outdoors,
featuring abundant cherry blossom trees 
with delicate pink flowers and
falling petals. A large body of 
reflective water, such as a lake or 
river, is in the foreground, with a strong
focus on the clear reflections of the 
surrounding scene and sky on its surface.
A picturesque wooden bridge with a railing 
spans the water or crosses a path nearby. 
A path leads through lush green grass and 
bushes along the riverbank or shore. 
A dense forest covers rolling hills, 
and majestic mountains form a dramatic
mountainous horizon in the background 
under a cloudy sky. No humans are presented.
best quality, highly detailed.
\end{verbatim}
The model generated a natural image, demonstrating the capability of the trained architecture.

\subsection{Evaluation Metrics}
\subsubsection{FID (Fréchet Inception Distance)}
FID \cite{heusel2017gans} measures the similarity between the feature distributions of generated and real images. Features are extracted using a pre-trained Inception model, and the Fréchet distance between the two (modeled as multivariate Gaussian) distributions is calculated. A lower FID score indicates that the generated image distribution is closer to the real distribution, signifying higher quality and diversity.

\subsubsection{CLIP Score (CLIP Similarity)}
CLIP Score \cite{hessel2021clipscore} evaluates the semantic similarity between a generated image and its corresponding text prompt, using the CLIP model. A higher score indicates better alignment between the image and the text.

\subsubsection{GenEval (Generative Evaluation)}
GenEval \cite{ghosh2023geneval} refers to a framework for evaluating T2I alignment, often focusing on object-level metrics. The specific calculation depends on the context.

\subsubsection{TIFA Score (Text-to-Image Fidelity and Alignment)}
TIFA Score \cite{hu2023tifa} evaluates T2I faithfulness by posing visual questions about the generated image based on the prompt. It assesses whether objects, attributes, and relationships are correctly rendered. A higher score indicates better prompt fidelity. This metric is critical for our use case, as high fidelity reduces the need for ``prompt fishing" (generating many samples to find one correct one), thereby saving time in a production environment.

\subsubsection{Aesthetic Score}
Aesthetic Score \cite{lee2019image} estimates the visual appeal of an image. It is typically predicted by a model trained on human preference data. A higher score suggests the image is more likely to be perceived as visually pleasing.

\subsubsection{Win Rate}
Win Rate measures human preference. Raters are shown images from different models and asked to vote for the superior one. A higher win rate indicates a model's outputs are preferred over its competitors. In our evaluation, raters were shown the prompt and the image to assess both quality and prompt fidelity.

\subsection{Quantitative Evaluation and Comparison with Other Models}
\subsubsection{Various Evaluation Metrics}
Table \ref{tab:metrics} shows a quantitative comparison against SD1.5 and SDXL.

\begin{table}[htbp]
\centering
\caption{Evaluation Metric Comparison}
\label{tab:metrics}
\begin{tabular}{@{}lcccc@{}}
\toprule
Metric & Direction & SD 1.5 & SDXL & oboro: \\ \midrule
FID & $\downarrow$ & 19.6 & 24.3 & 22.7 \\
CLIP Score & $\uparrow$ & 0.31 & 0.31 & 0.31 \\
GenEval & $\uparrow$ & 42\% & 54\% & 30\% \\
TIFA Score & $\uparrow$ & 0.52 & 0.62 & 0.85 \\
Aesthetic Score & $\uparrow$ & 4.8 & 6.4 & 4.5 \\
Win Rate & $\uparrow$ & - & 11.3\%* & 88.3\%* \\ \bottomrule
\end{tabular}
\begin{flushleft}
The $\uparrow$$\downarrow$ for each metric indicates whether $\uparrow$ (higher) is better or $\downarrow$ (lower) is better.
* indicates the comparison result between oboro:anime and SDXL.
\end{flushleft}
\end{table}

\subsubsection{Learning Time \& Dataset}
\begin{figure}[htbp]
  \centering
  \includegraphics[width=8cm]{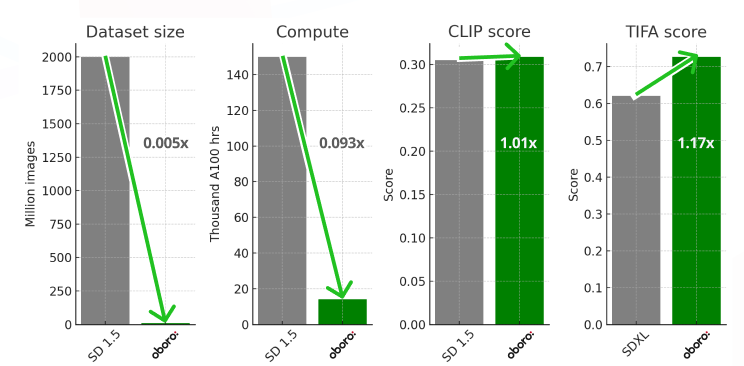}
  \caption{Development Results: Achieving Equivalent Levels with Limited Data. This shows \textit{oboro} achieving comparable performance to SD 1.5 and SDXL with significantly less data and compute.}
  \label{fig:comparison_slide}
\end{figure}

Our model achieved evaluation metrics comparable to SD1.5 and SDXL, but with significantly reduced training time and data. The model achieved this performance using approximately 1/200th of the dataset size and 1/10th of the computational time associated with the larger models. This validates our primary objective: developing a model capable of high-performance training on small, copyright-cleared datasets. (See Fig. \ref{fig:comparison_slide}).

\section{Results and Discussion}
\subsection{Summary of Experimental Results}
We developed two models: a foundation model (``oboro:base") and a specialized model (``oboro:anime"). By adopting a new architecture, we successfully reduced the computational resources and dataset size required for training. The resulting image quality is comparable to competing models, achieving this level with 1/10th the training time and 1/50th the image count. As copyright-cleared datasets are inherently small, the ability to achieve competitive quality with 1/50th the data is a distinct advantage.

``oboro:base" serves as the foundation model prior to specialized training. ``oboro:anime" is a test model trained to simulate specialization for an anime studio. While we originally planned to train ``oboro:anime" on data provided by anime studios, we pivoted to using license-verified illustration data to allow more time for stakeholder consensus and rights-framework finalization. To ensure strict compliance with rights agreements and confidentiality, ``oboro:anime" is not publicly released.

\subsection{Model Characteristics and Limitations}
The primary strength of this model is its rendering of light and shadow (the namesake of ``oboro"), enabling the generation of images with effective contrast. Color and contrast quality are also good. The use of the T5-XXL text encoder provides high prompt responsiveness, making it easier to generate intended images.

However, several limitations were identified. The model exhibits a tendency to over-render details, sometimes resulting in a grainy texture across the image, potentially due to overfitting. The model is not proficient at rendering human figures; stable quality in this domain remains a future challenge. Additionally, due to the dataset's composition, the model is constrained from generating images in certain genres, such as fantasy or specific illustration styles.

These limitations may not be critical for a foundation model, as the intended workflow involves specialized fine-tuning by anime studios on their proprietary datasets. This allows for the addition of specific characters or fantasy elements. Indeed, improvements in character generation were observed during the ``oboro:anime" training tests.

\subsection{Future Issues and Prospects}
We have developed a foundation model and a specialized model test case. We are also developing auxiliary technologies such as LoRA and ControlNet, as well as related application tools. We plan to leverage the findings from this project for future model development.

\section{Model Release and Usage}
\subsection{Public Repository (Model Data/Inference Code)}
The model data and inference code are available at:
\url{https://huggingface.co/aihub-geniac/oboro}

\subsection{Setup and Execution Method}
\subsubsection{Agree to the FLUX.1 [schnell] License}
This model uses the VAE of FLUX.1 [schnell] \cite{bfl2025flux} to restore images from latent space. To use the image encoder/decoder, please open the page \url{https://huggingface.co/black-forest-labs/FLUX.1-schnell} and agree to the license. The files will be automatically downloaded when the image generation is executed, so manual download is not necessary.

\subsubsection{Download aihub-geniac/oboro files}
Please download all files included in \url{https://huggingface.co/aihub-geniac/oboro}.

\subsubsection{Install Dependencies}
Install the dependent libraries. Please execute in an appropriate virtual environment as needed.
\begin{verbatim}
pip install -r requirements.txt
\end{verbatim}

\subsubsection{Login to Hugging Face CLI}
Log in in advance using the Hugging Face CLI to download the necessary models. After executing the command, you will be prompted to enter your Hugging Face account token (read-only permissions are fine), so please enter it.
\begin{verbatim}
huggingface-cli login
\end{verbatim}

\subsubsection{Run Image Generation}
\small{
\begin{verbatim}
python src/infer.py --prompts 'A serene ...'
  --image_size '[416,736]' --cfg_scale '5.0'
  --model_path 'oboro-base-v1-1b.safetensors'
  --output_dir 'output'
\end{verbatim}
}
(Note: Full prompt text is omitted for brevity, see Section V-A).

\subsection{License}
The oboro:base pre-trained model and inference code are released under the Apache License, Version 2.0.

\section*{Acknowledgments}
This project was realized with the assistance of computational resources under ``GENIAC," Japan's generative AI development capability enhancement project led by the Ministry of Economy, Trade and Industry (METI) and the New Energy and Industrial Technology Development Organization (NEDO). We express our deepest gratitude to all involved parties.

\bibliographystyle{IEEEtran}
\bibliography{oboro-tech-paper-11}

@string{IEEE_J_CVPR = {Proc. IEEE/CVF Conf. Comput. Vis. Pattern Recognit. (CVPR)}}

@string{IEEE_J_ICCV = {Proc. IEEE/CVF Int. Conf. Comput. Vis. (ICCV)}}

@string{IEEE_J_NIPS = {Adv. Neural Inf. Process. Syst. (NIPS)}}

@string{IEEE_J_ICLR = {Proc. Int. Conf. Learn. Represent. (ICLR)}}

@inproceedings{rombach2022highres,
  author    = {Rombach, Robin and Blattmann, Andreas and Lorenz, Dominik and Esser, Patrick and Ommer, Bj\"{o}rn},
  title     = {High-Resolution Image Synthesis with Latent Diffusion Models},
  booktitle = IEEE_J_CVPR,
  month     = {June},
  year      = {2022},
  pages     = {10684--10695},
  doi       = {10.1109/CVPR52688.2022.01042}
}

@misc{zhang2023adding,
  author       = {Zhang, Lvmin and Rao, Anyi and Agrawala, Maneesh},
  title        = {Adding Conditional Control to Text-to-Image Diffusion Models},
  year         = {2023},
  month        = {Feb},
  eprint       = {2302.05543},
  archiveprefix = {arXiv},
  primaryclass = {cs.CV}
}

@misc{hu2021lora,
  author       = {Hu, Edward J. and Shen, Yelong and Wallis, Phillip and Allen-Zhu, Zeyuan and Li, Yuanzhi and Wang, Shean and Wang, Lu and Chen, Weizhu},
  title        = {LoRA: Low-Rank Adaptation of Large Language Models},
  year         = {2021},
  month        = {June},
  eprint       = {2106.09685},
  archiveprefix = {arXiv},
  primaryclass = {cs.CL}
}

@misc{peebles2022scalable,
  author       = {Peebles, William and Xie, Saining},
  title        = {Scalable Diffusion Models with Transformers},
  year         = {2022},
  month        = {Dec},
  eprint       = {2212.09748},
  archiveprefix = {arXiv},
  primaryclass = {cs.CV}
}

@misc{stability2024sd3,
  author = {{Stability AI}},
  title  = {Stable Diffusion 3},
  year   = {2025},
  month  = {Aug},
  note   = {Accessed: Aug. 5, 2025. (Note: Date follows original JP report context)},
  howpublished = {\url{https://stability.ai/news/stable-diffusion-3}}
}

@misc{schuhmann2022laion5b,
  author = {Schuhmann, Christoph and others},
  title  = {LAION-5B: An open large-scale dataset for training next generation image-text models},
  year   = {2022},
  note   = {Blog post},
  howpublished = {\url{https://laion.ai/blog/laion-5b/}}
}

@misc{goodfellow2014gan,
  author       = {Goodfellow, Ian J. and Pouget-Abadie, Jean and Mirza, Mehdi and Xu, Bing and Warde-Farley, David and Ozair, Sherjil and Courville, Aaron and Bengio, Yoshua},
  title        = {Generative Adversarial Networks},
  year         = {2014},
  month        = {June},
  eprint       = {1406.2661},
  archiveprefix = {arXiv},
  primaryclass = {cs.LG}
}

@misc{ho2020ddpm,
  author       = {Ho, Jonathan and Jain, Ajay and Abbeel, Pieter},
  title        = {Denoising Diffusion Probabilistic Models},
  year         = {2020},
  month        = {June},
  eprint       = {2006.11239},
  archiveprefix = {arXiv},
  primaryclass = {cs.LG}
}

@misc{ramesh2021zero,
  author       = {Ramesh, Aditya and Pavlov, Mikhail and Goh, Gabriel and Gray, Scott and Voss, Chelsea and Radford, Alec and Chen, Mark and Sutskever, Ilya},
  title        = {Zero-Shot Text-to-Image Generation},
  year         = {2021},
  month        = {Feb},
  eprint       = {2102.12092},
  archiveprefix = {arXiv},
  primaryclass = {cs.CV}
}

@misc{radford2021learning,
  author       = {Radford, Alec and Kim, Jong Wook and Hallacy, Chris and Ramesh, Aditya and Goh, Gabriel and Agarwal, Sandhini and Sastry, Girish and Askell, Amanda and Mishkin, Pamela and Clark, Jack and Krueger, Gretchen and Sutskever, Ilya},
  title        = {Learning Transferable Visual Models From Natural Language Supervision},
  year         = {2021},
  month        = {Feb},
  eprint       = {2103.00020},
  archiveprefix = {arXiv},
  primaryclass = {cs.CV}
}

@misc{raffel2019exploring,
  author       = {Raffel, Colin and Shazeer, Noam and Roberts, Adam and Lee, Katherine and Narang, Sharan and Matena, Michael and Zhou, Yanqi and Li, Wei and Liu, Peter J.},
  title        = {Exploring the Limits of Transfer Learning with a Unified Text-to-Text Transformer},
  year         = {2019},
  month        = {Oct},
  eprint       = {1910.10683},
  archiveprefix = {arXiv},
  primaryclass = {cs.LG}
}

@misc{c4dataset,
  author = {{TensorFlow Datasets}},
  title  = {C4 (Colossal Clean Crawled Corpus)},
  year   = {2019},
  howpublished = {\url{https://www.tensorflow.org/datasets/catalog/c4}}
}

@misc{gemma2024,
  author = {{Google DeepMind}},
  title  = {Gemma: Open Models Based on Gemini Research and Technology},
  year   = {2024},
  howpublished = {\url{https://deepmind.google/models/gemma/}}
}

@misc{qin2025lumina,
  author       = {Qin, Qi and Zhuo, Le and Xin, Yi and Du, Ruoyi and Li, Zhen and Fu, Bin and Lu, Yiting and Yuan, Jiakang and Li, Xinyue and Liu, Dongyang and Zhu, Xiangyang and Zhang, Manyuan and Beddow, Will and Millon, Erwann and Perez, Victor and Wang, Wenhai and He, Conghui and Zhang, Bo and Liu, Xiaohong and Li, Hongsheng and Qiao, Yu and Xu, Chang and Gao, Peng},
  title        = {Lumina-Image 2.0: A Unified and Efficient Image Generative Framework},
  year         = {2025},
  month        = {Mar},
  eprint       = {2503.21758},
  archiveprefix = {arXiv},
  primaryclass = {cs.CV}
}

@misc{ronneberger2015unet,
  author       = {Ronneberger, Olaf and Fischer, Philipp and Brox, Thomas},
  title        = {U-Net: Convolutional Networks for Biomedical Image Segmentation},
  year         = {2015},
  month        = {May},
  eprint       = {1505.04597},
  archiveprefix = {arXiv},
  primaryclass = {cs.CV}
}

@misc{auradiffusion2024vae,
  author = {{AuraDiffusion}},
  title  = {16ch-vae},
  year   = {2024},
  howpublished = {\url{https://huggingface.co/AuraDiffusion/16ch-vae}}
}

@misc{clipvitlarge,
  author = {{OpenAI}},
  title  = {CLIP Model Card: clip-vit-large-patch14},
  year   = {2021},
  howpublished = {\url{https://huggingface.co/openai/clip-vit-large-patch14}}
}

@misc{podell2023sdxl,
  author       = {Podell, Dustin and English, Zion and Lacey, Kyle and Blattmann, Andreas and Dockhorn, Tim and Müller, Jonas and Penna, Joe and Rombach, Robin},
  title        = {SDXL: Improving Latent Diffusion Models for High-Resolution Image Synthesis},
  year         = {2023},
  month        = {July},
  eprint       = {2307.01952},
  archiveprefix = {arXiv},
  primaryclass = {cs.CV}
}

@misc{openclipvitbigG,
  author = {{LAION}},
  title  = {CLIP-ViT-bigG-14-laion2B-39B-b160k},
  year   = {2023},
  howpublished = {\url{https://huggingface.co/laion/CLIP-ViT-bigG-14-laion2B-39B-b160k}}
}

@misc{bfl2025flux,
  author       = {{Black Forest Labs} and Batifol, Stephen and Blattmann, Andreas and Boesel, Frederic and Consul, Saksham and Diagne, Cyril and Dockhorn, Tim and English, Jack and English, Zion and Esser, Patrick and Kulal, Sumith and Lacey, Kyle and Levi, Yam and Li, Cheng and Lorenz, Dominik and Müller, Jonas and Podell, Dustin and Rombach, Robin and Saini, Harry and Sauer, Axel and Smith, Luke},
  title        = {FLUX.1 Kontext: Flow Matching for In-Context Image Generation and Editing in Latent Space},
  year         = {2025},
  month        = {June},
  eprint       = {2506.15742},
  archiveprefix = {arXiv},
  primaryclass = {cs.CV}
}

@misc{zhuo2024lumina,
  author       = {Zhuo, Le and Du, Ruoyi and Xiao, Han and Li, Yangguang and Liu, Dongyang and Huang, Rongjie and Liu, Wenze and Zhao, Lirui and Wang, Fu-Yun and Ma, Zhanyu and Luo, Xu and Wang, Zehan and Zhang, Kaipeng and Zhu, Xiangyang and Liu, Si and Yue, Xiangyu and Liu, Dingning and Ouyang, Wanli and Liu, Ziwei and Qiao, Yu and Li, Hongsheng and Gao, Peng},
  title        = {Lumina-Next: Making Lumina-T2X Stronger and Faster with Next-DiT},
  year         = {2024},
  month        = {June},
  eprint       = {2406.18583},
  archiveprefix = {arXiv},
  primaryclass = {cs.CV}
}

@misc{shazeer2020glu,
  author       = {Shazeer, Noam},
  title        = {GLU Variants Improve Transformer},
  year         = {2020},
  month        = {Feb},
  eprint       = {2002.05202},
  archiveprefix = {arXiv},
  primaryclass = {cs.LG}
}

@misc{heo2024rope,
  author       = {Heo, Byeongho and Park, Song and Han, Dongyoon and Yun, Sangdoo},
  title        = {Rotary Position Embedding for Vision Transformer},
  year         = {2024},
  month        = {July},
  eprint       = {2403.13298},
  archiveprefix = {arXiv},
  primaryclass = {cs.CV}
}

@misc{defazio2024road,
  author       = {Defazio, Aaron and Yang, Xingyu Alice and Mehta, Harsh and Mishchenko, Konstantin and Khaled, Ahmed and Cutkosky, Ashok},
  title        = {The Road Less Scheduled},
  year         = {2024},
  month        = {May},
  eprint       = {2405.15682},
  archiveprefix = {arXiv},
  primaryclass = {cs.LG}
}

@misc{mehta2023matchatts,
  author        = {Mehta, Shivam and Tu, Ruibo and Beskow, Jonas and Sz{\'e}kely, {\'E}va and Henter, Gustav Eje},
  title         = {{Matcha-TTS}: A fast {TTS} architecture with conditional flow matching},
  year          = {2023},
  month         = {Sep},
  eprint        = {2309.03199},
  archiveprefix = {{arXiv}},
  primaryclass  = {{cs.SD}}
}

@misc{liu2022flow,
  author       = {Liu, Xingchao and Gong, Chengyue and Liu, Qiang},
  title        = {Flow Straight and Fast: Learning to Generate and Transfer Data with Rectified Flow},
  year         = {2022},
  month        = {Sep},
  eprint       = {2209.03003},
  archiveprefix = {arXiv},
  primaryclass = {cs.LG}
}

@misc{tong2023improving,
  author       = {Tong, Alexander and Fatras, Kilian and Malkin, Nikolay and Huguet, Guillaume and Zhang, Yanlei and Rector-Brooks, Jarrid and Wolf, Guy and Bengio, Yoshua},
  title        = {Improving and generalizing flow-based generative models with minibatch optimal transport},
  year         = {2023},
  month        = {Feb},
  eprint       = {2302.00482},
  archiveprefix = {arXiv},
  primaryclass = {cs.LG}
}

@misc{t5v11xxl,
  author = {{Google}},
  title  = {T5-v1\_1-xxl Model Card},
  year   = {2020},
  howpublished = {\url{https://huggingface.co/google/t5-v1_1-xxl}}
}

@misc{megalith10m,
  author = {{Ollin}},
  title  = {Megalith-10m Dataset},
  year   = {2024},
  howpublished = {\url{https://huggingface.co/datasets/madebyollin/megalith-10m}}
}

@misc{sharecaptioner,
  author = {Lin-Chen},
  title  = {ShareCaptioner Model Card},
  year   = {2024},
  howpublished = {\url{https://huggingface.co/Lin-Chen/ShareCaptioner}}
}

@misc{florence2,
  author = {{Microsoft}},
  title  = {Florence-2-large Model Card},
  year   = {2024},
  howpublished = {\url{https://huggingface.co/microsoft/Florence-2-large}}
}

@misc{aestheticpredictor,
  author = {discus0434},
  title  = {aesthetic-predictor-v2-5},
  year   = {2023},
  howpublished = {\url{https://github.com/discus0434/aesthetic-predictor-v2-5}}
}

@misc{pfntechblog2024,
  author = {{Preferred Networks, Inc.}},
  title  = {Inference and visualization of segmentation results using trained models},
  year   = {2024},
  month  = {Apr},
  note   = {(Original title in Japanese)},
  howpublished = {\url{https://tech.preferred.jp/ja/blog/inference-for-data-preprocessing/}}
}

@misc{shi2016realtime,
  author       = {Shi, Wenzhe and Caballero, Jose and Huszár, Ferenc and Totz, Johannes and Aitken, Andrew P. and Bishop, Rob and Rueckert, Daniel and Wang, Zehan},
  title        = {Real-Time Single Image and Video Super-Resolution Using an Efficient Sub-Pixel Convolutional Neural Network},
  year         = {2016},
  month        = {Sep},
  eprint       = {1609.05158},
  archiveprefix = {arXiv},
  primaryclass = {cs.CV}
}

@misc{loshchilov2017decoupled,
  author       = {Loshchilov, Ilya and Hutter, Frank},
  title        = {Decoupled Weight Decay Regularization},
  year         = {2017},
  month        = {Nov},
  eprint       = {1711.05101},
  archiveprefix = {arXiv},
  primaryclass = {cs.LG}
}

@inproceedings{heusel2017gans,
  author    = {Heusel, Martin and Ramsauer, Hubert and Unterthiner, Thomas and Nessler, Bernhard and Hochreiter, Sepp},
  title     = {{GANs} Trained by a Two Time-Scale Update Rule Converge to a Local Nash Equilibrium},
  booktitle = IEEE_J_NIPS,
  year      = {2017},
  pages     = {6629--6640}
}

@misc{hessel2021clipscore,
  author       = {Hessel, Jack and Holtzman, Ari and Forbes, Maxwell and Le Bras, Ronan and Choi, Yejin},
  title        = {CLIPScore: A Reference-free Evaluation Metric for Image Captioning},
  year         = {2021},
  month        = {Apr},
  eprint       = {2104.08718},
  archiveprefix = {arXiv},
  primaryclass = {cs.CL}
}

@misc{ghosh2023geneval,
  author       = {Ghosh, Dhruba and Hajishirzi, Hanna and Schmidt, Ludwig},
  title        = {GenEval: An Object-Focused Framework for Evaluating Text-to-Image Alignment},
  year         = {2023},
  month        = {Oct},
  eprint       = {2310.11513},
  archiveprefix = {arXiv},
  primaryclass = {cs.CV}
}

@misc{hu2023tifa,
  author       = {Hu, Yushi and Liu, Benlin and Kasai, Jungo and Wang, Yizhong and Ostendorf, Mari and Krishna, Ranjay and Smith, Noah A.},
  title        = {TIFA: Accurate and Interpretable Text-to-Image Faithfulness Evaluation with Question Answering},
  year         = {2023},
  month        = {Mar},
  eprint       = {2303.11897},
  archiveprefix = {arXiv},
  primaryclass = {cs.CV}
}

@inproceedings{lee2019image,
  author    = {Lee, Jun-Tae and Kim, Chang-Su},
  title     = {Image Aesthetic Assessment Based on Pairwise Comparison - A Unified Approach to Score Regression, Binary Classification, and Personalization},
  booktitle = IEEE_J_ICCV,
  year      = {2019},
  pages     = {1191--1200},
  doi       = {10.1109/ICCV.2019.00128}
}

@misc{stability2024sd3paper,
  author = {{Stability AI}},
  title  = {Stable Diffusion 3: Research Paper},
  year   = {2024},
  month  = {Mar},
  howpublished = {\url{https://stability.ai/news/stable-diffusion-3-research-paper}}
}

@misc{stability2024sd35,
  author = {{Stability AI}},
  title  = {Introducing Stable Diffusion 3.5},
  year   = {2024},
  month  = {Oct},
  howpublished = {\url{https://stability.ai/news/introducing-stable-diffusion-3-5}}
}

@misc{su2021roformer,
  author       = {Su, Jianlin and Lu, Yu and Pan, Shengfeng and Wen, Bo and Liu, Yunfeng},
  title        = {RoFormer: Enhanced Transformer with Rotary Position Embedding},
  year         = {2021},
  month        = {Apr},
  eprint       = {2104.09864},
  archiveprefix = {arXiv},
  primaryclass = {cs.CL}
}

@misc{shleifer2021normformer,
  author       = {Shleifer, Sam and Weston, Jason and Ott, Myle},
  title        = {NormFormer: Improved Transformer Pretraining with Extra Normalization},
  year         = {2021},
  month        = {Oct},
  eprint       = {2110.09456},
  archiveprefix = {arXiv},
  primaryclass = {cs.CL}
}

@misc{zhai2023siglip,
  author       = {Zhai, Xiaohua and Kolesnikov, Alexander and Mustafa, Basil and Beyer, Lucas},
  title        = {Sigmoid Loss for Language-Image Pre-Training (SigLIP)},
  year         = {2023},
  month        = {Mar},
  eprint       = {2303.15343},
  archiveprefix = {arXiv},
  primaryclass = {cs.CV}
}

@inproceedings{wang2023qknorm,
  author    = {Wang, Chengnan and Bai, Yunsheng and Zhai, Bohan and Chen, Cen},
  title     = {{QK-Norm}: Better Transformer Training with Query-Key Normalization},
  booktitle = IEEE_J_ICLR,
  year      = {2023},
  url       = {https://openreview.net/forum?id=E4jFcMHgCQa}
}

@misc{rye2025,
  author = {{Astral}},
  title  = {Rye - A Python Toolchain Manager},
  year   = {2025},
  howpublished = {\url{https://rye.astral.sh/}}
}

@misc{uv2025,
  author = {{Astral}},
  title  = {uv - An Extremely Fast Python Package Installer and Resolver},
  year   = {2025},
  howpublished = {\url{https://docs.astral.sh/uv/}}
}

@inproceedings{ding2021cogview,
  author    = {Ding, Ming and Yang, Zhuoyi and Hong, Wenyi and Zheng, Wendi and Zhou, Chang and Yin, Da and Lin, Junyang and Zou, Xu and Shao, Zhou and Yang, Hongxia and others},
  title     = {Cogview: Mastering text-to-image generation via transformers},
  booktitle = IEEE_J_NIPS,
  volume    = {34},
  year      = {2021},
  pages     = {19822--19835}
}

@misc{egov2018copyright,
  author = {{e-Gov Japan (Digital Agency)}},
  title  = {Copyright Act Article 30-4 (Reproduction, etc. for information analysis)},
  year   = {2018},
  howpublished = {\url{https://elaws.e-gov.go.jp/document?lawid=345AC0000000048}}
}

@misc{bunka2024ai,
  author = {{Agency for Cultural Affairs, Japan}},
  title  = {About AI and Copyright (View on Generative AI)},
  year   = {2024},
  howpublished = {\url{https://www.bunka.go.jp/seisaku/bunkashingikai/chosakuken/houkokusho/94138001.html}}
}

@misc{bfl2024announcing,
  author = {{Black Forest Labs}},
  title  = {Announcing Black Forest Labs},
  year   = {2024},
  month  = {Aug},
  howpublished = {\url{https://bfl.ai/blog/24-08-01-bfl}}
}

@misc{zhang2025encoder,
  author       = {Zhang, Biao and Moiseev, Fedor and Ainslie, Joshua and Suganthan, Paul and Ma, Min and Bhupatiraju, Surya and Lebron, Fede and Firat, Orhan and Joulin, Armand and Dong, Zhe},
  title        = {Encoder-Decoder Gemma: Improving the Quality-Efficiency Trade-Off via Adaptation},
  year         = {2025},
  month        = {Apr},
  eprint       = {2504.06225},
  archiveprefix = {arXiv},
  primaryclass = {cs.CL}
}

@misc{vaswani2017attention,
  author       = {Vaswani, Ashish and Shazeer, Noam and Parmar, Niki and Uszkoreit, Jakob and Jones, Llion and Gomez, Aidan N. and Kaiser, Lukasz and Polosukhin, Illia},
  title        = {Attention Is All You Need},
  year         = {2017},
  month        = {June},
  eprint       = {1706.03762},
  archiveprefix = {arXiv},
  primaryclass = {cs.CL}
}

@misc{mochi2024,
  author       = {Zeng, Anda and others},
  title        = {Mochi: A 20B-parameter Text-to-Image Model with Unmatched Comp-Gen Ability},
  year         = {2024},
  eprint       = {2407.08697},
  archiveprefix = {arXiv}
}

@misc{kumar2025zclip,
      title={ZClip: Adaptive Spike Mitigation for LLM Pre-Training},
      author={Abhay Kumar and Louis Owen and Nilabhra Roy Chowdhury and Fabian Güra},
      year={2025},
      eprint={2504.02507},
      archivePrefix={arXiv},
      primaryClass={cs.LG},
      url={https://arxiv.org/abs/2504.02507},
}

\end{document}